\documentclass[journal]{IEEEtran}

\usepackage{booktabs}       
\usepackage{amsfonts}       
\usepackage{nicefrac}       
\usepackage{microtype}      
\usepackage{color}
\usepackage{amsmath}
\usepackage{multirow}
\usepackage{diagbox}

\usepackage{cite}
\usepackage{amsmath,amssymb,amsfonts}
\usepackage{graphicx}
\usepackage{textcomp}
\usepackage{xcolor}
\usepackage{url}

\usepackage[font=footnotesize,labelfont=bf]{caption}

\usepackage{cite}
\usepackage{amsmath,amssymb,amsfonts}

\usepackage{graphicx}

\ifCLASSINFOpdf
\else
\fi
\begin{document}

%
\title{BCNN: Binary Complex Neural Network}
%
%
%

\author{\IEEEauthorblockN{Yanfei~Li$^{\dag}$, Tong~Geng$^{\ddag}$, Ang~Li$^{\ddag}$, and~Huimin~Yu$^{\dag}$}\\
\IEEEauthorblockA{$^\dag$Zhejiang University, Hangzhou, Zhejiang, China\\ $^\ddag$ Pacific Northwest National Laboratory, Richland, WA, USA  \\
aoxue18@zju.edu.cn, tong.geng@pnnl.gov, ang.li@pnnl.gov, yhm2005@zju.edu.cn}
}

\maketitle

\begin{abstract}
Binarized neural networks, or BNNs, show great promise in edge-side applications with resource limited hardware, but raise the concerns of reduced accuracy. Motivated by the complex neural networks, in this paper we introduce complex representation into the BNNs and propose Binary complex neural network -- a novel network design that processes binary complex inputs and weights through complex convolution, but still can harvest the extraordinary computation efficiency of BNNs. To ensure fast convergence rate, we propose novel BCNN based batch normalization function and weight initialization function. Experimental results on Cifar10 and ImageNet using state-of-the-art network models (e.g., ResNet, ResNetE and NIN) show that BCNN can achieve better accuracy compared to the original BNN models. BCNN improves BNN by strengthening its learning capability through complex representation and extending its applicability to complex-valued input data. The source code of BCNN will be released on GitHub.

\end{abstract}


%
\IEEEpeerreviewmaketitle

\section{Introduction}


Deep neural networks (DNNs) recently achieved tremendous success in many computer vision applications. Aiming at replicating similar success for practical edge-side utilization, researchers are tweaking the DNN models for resources limited hardware. 
Binary neural networks (BNNs) \cite{courbariaux2016binarized, hubara2016binarized}, which adopt only a bit for a neuron, stand out as one of the most promising approaches. To accommodate constrained hardware budget, rather than deducting the number of neurons in a model, BNN reduces the number of bits per neuron to the extreme --- each element of the input, weight and activation of a BNN layer is merely a single binary value, implying +1 or -1.

BNNs have demonstrated several appealing features for embedded utilization: (i) \emph{Execution efficiency:} the natural mapping from a BNN neuron to a digital bit makes BNN extremely hardware-friendly. From the computation perspective, each 32 or 64 full-precision dot-product calculations can be aggregated into a Boolean exclusive-or plus a population-count operation \cite{courbariaux2016binarized}, improving the execution efficiency by more than 10x \cite{li2019bstc}. From the memory perspective, rather than using a 32-bit single-precision floating-point, or a 16-bit half-precision floating-point, in BNNs each neuron uses one bit, substantially improving the utilization of the memory storage and bandwidth \cite{li2020accelerating}. The combination of the two effects can bring over three orders of latency reduction for single image inference compared to full-precision DNNs on GPUs \cite{li2019bstc}. (ii) \emph{Low-cost:} Due to simpler hardware logic (e.g., avoiding using floating-point multiplier) and diminished memory demand, the hardware cost for implementing BNNs is much lower than DNNs \cite{geng2019o3bnn, geng2020o3bnn}. (iii) \emph{Energy-efficiency:} Due to low-cost hardware operations and smaller chip-area \cite{geng2019lp}, BNN-based designs are very friendly to portable devices with limited cycle life of batteries.  
(iv) \emph{Robustness}: Due to the discrete parameter space through the binarization of weight, BNN shows better robustness than normal DNNs \cite{galloway2017attacking}, while certain properties can be formally verified \cite{narodytska2018formal, narodytska2018verifying}. Because of these advantages, BNNs have been utilized for a variety of practical applications, such as auto-driving \cite{chen2020gpu}, COVID face-cover detection \cite{fasfous2021binarycop}, smart agriculture \cite{huang2021fpga}, image enhancement \cite{ma2019efficient}, 3D objection detection \cite{ma2018binary}, etc. 
 
Although BNNs exhibit these attractive features, concerns have been raised over the reduced accuracy compared to DNNs, which is largely due to the loss of information in the binarization process, and the reduced model capacity. Ever since the proposal of BNNs, continuous efforts have been invested from the machine learning community on improving BNN accuracy  \cite{rastegari2016xnor,zhou2016dorefa,tang2017train,lin2017towards,darabi2018bnn,ghasemzadeh2018rebnet,zhuang2019structured,zhu2019binary,bethge2019binarydensenet,bethge2021meliusnet}, as briefly summarized in the next section. 

In the meanwhile, complex-valued neural networks \cite{trabelsi2017deep} have been proposed as an amendment to the normal DNNs. Most existing DNNs adopt real-valued representation for the inputs and weights. However, considering the richer representational capacity \cite{wisdom2016full}, the better generalization capability \cite{hirose2012generalization}, and the potential to facilitate noise-robust memory retrieval  \cite{danihelka2016associative}, deep complex networks have been formulated \cite{trabelsi2017deep}, in which the inputs, the weights and the outputs are all complex values. Correspondingly, the convolution, the batch normalization, the activation, etc. are reformed in complex operations. It has been shown that complex networks can deliver comparable or even better accuracy than DNNs under the same model capacity. A particularly attractive feature of the complex network is the ability to embed phase information of the input data naturally into the network representation. The phase information is critical for deterministic signals, such as neuronal rhythms in the brain \cite{reichert2013neuronal}, Polarimetric synthetic aperture radar (PolSAR) images \cite{cao2019pixel}, Fourier representation of wave, speech data \cite{choi2018phase}, multi-channel images like MRI \cite{cole2020analysis}, etc.

Considering the adoption of complex networks for terminal scenarios, in this work, we propose a novel network called \emph{Binary Complex Neural Network} (BCNN) that integrates BNNs and complex neural networks.
BCNN extends BNN with its richer representation capacity of complex numbers, but still conserving the high computation efficiency of BNNs for resource limited hardware. In BCNN, the input, weight and output of a layer are all binary complex values, i.e., one of \{$1+i$, $1-i$, $-1+i$, $-1-i$\}, using \textbf{dual-bits} per neuron --- one for the real part and one for the imaginary part. We propose binary complex convolution, which follows the complex number computation rules, but can still be calculated through the assembly-level \emph{xnor-popcnt} machine instructions available in most hardware. Tackling the expensive computation cost of the original complex network in batch normalization \cite{trabelsi2017deep}, which involves matrix inversion and square-rooting, in this work we propose a new batch normalization approach that can significantly simplify the computation logic while facilitate the convergence of the training. Furthermore, we propose a BCNN weight initialization strategy to accelerate the convergence speed and mitigate the chances of gradient explosion/vanishing. Evaluation results on the Cifar10 and ImageNet datasets show that BCNN can achieve better accuracy than BNNs  using state-of-the-art models like ResNet \cite{he2016deep}, ResNetE \cite{liu2018bi, bethge2018training}, and NIN \cite{lin2013network}.
Our contribution in this paper are:
\begin{enumerate}
\item We propose the concept of binary complex number, including its dual-bit storage format and the binary complex computation mechanism. 
\item We propose the binary complex neural network (BCNN), including quadrant binarization and its gradient.
\item We propose a BCNN-oriented batch normalization function, significantly lowering the computation cost.
\item We propose a BCNN-oriented weight initialization function, facilitating better convergence for the training.
\item Evaluations on Cifar10 and ImageNet datasets show that BCNN can achieve better accuracy than original BNNs.
\end{enumerate}

This paper is organized as follows. We summarize existing literature in Section-II, covering theoretical research about BNNs, the major approaches to enhance BNN accuracy, and the complex neural networks. We present the design details of BCNN in Section-III, covering the definition of binary complex numbers (Section-III-A), the quadrant binarization function (Section-III-B), the batch-normalization (Section-III-C), and the weight initialization (Section-III-D). We evaluate BCNN in Section-IV and conclude in Section-V.


\section{Related Work}
We briefly discuss existing literature about BNNs and complex neural networks.

\subsection{Binarized Neural Networks}
Binarized Neural Network (BNN) was originally evolved from \emph{Binarized Weights Network} (BWN) \cite{courbariaux2015binaryconnect} in which only weights are binarized. The foundation of modern BNNs were laid by the two cornerstone works \cite{courbariaux2016binarized, hubara2016binarized} in which the fundamental components of BNNs were proposed, including (1) the binarization function and its approximated gradient through \emph{straight-through estimator} (STE) on latent variables; (2) batch-normalization, which is crucial for BNNs to be able to converge; (3) the necessity to keep full-precision for the first and last layers. It was later explained by Anderson and Berg \cite{anderson2017high} on why BNNs could effectively approximate DNNs: (i) the binary vector through binarization preserves the direction of DNN real-valued vectors in the high-dimensional geometry space; (ii) the bit dot-product (\texttt{popc(xnor())}), through batch-normalization, preserves the property of original DNN dot-product; (iii) the real-value convolution for the first layer can embed input images into high-dimensional binary space, which can then be effectively handled by binary operations.

BNNs are generally criticized for reduced accuracy compared to their DNN counterparts because of (1) Information loss due to input binarization and binary activation; (2) Reduced model capacity due to weight binarization (1 bit per neuron); and (3) Unsatisfied network structure or training methodologies as existing popular models and training strategies were mainly designed for real-value DNNs. Correspondingly, existing works propose to enhance BNN training accuracy via: (1) \emph{Reducing information loss}. This can be achieved by adding gain terms (i.e., scaling factors) to better approximate DNN activation \cite{rastegari2016xnor}. Gain terms are extracted based on the statistics of the inputs \cite{zhou2016dorefa,tang2017train}, or gradually learned with the training \cite{lin2017towards,darabi2018bnn}; (2) \emph{Enhancing BNN model capacity}. This is done by using multiple BNN components in the network (e.g., \emph{BENN} \cite{zhu2019binary} and \emph{Group-Net} \cite{zhuang2019structured}), or using more bits for a neuron where each bit represents a basis. The basis can be fixed to powers-of-two (e.g.,1,2,4,8,...) \cite{zhou2016dorefa}, or adjustable as residual basis \cite{tang2017train,ghasemzadeh2018rebnet}, or even learned during training \cite{lin2017towards}; (3) \emph{Designing BNN-specific network structure and training methods}. As most existing network models were designed for DNNs, researchers started to design BNN-oriented network structures, these include \emph{ResnetE} and \emph{BinaryDenseNet} \cite{bethge2019binarydensenet} which adopted more shortcuts for reusing information to maintain rich information flow in the network, and \emph{MeliusNet} \cite{bethge2021meliusnet}, which conserved the mainstream information flow as full-precision in the first 256 channels, but using the two-block BNN structure (i.e., a dense block with an improvement block) for learning and attaching learned results in separated 64 channels. In this way, most information loss due to binarization could be avoided. Additionally, \emph{Bi-Real Net} redirects the information-rich real-valued activation before binarization to the next block through a shortcut \cite{liu2018bi}. 

Alternative works concentrated on improving BNN training methodology. For example, focusing on the \texttt{sign} activation function and the STE gradient estimator, Alizadeh et al. showed that adapting the learning rate using second-moment methods was crucial for the successful adoption of STE in BNN training, compared with other optimizers \cite{alizadeh2018empirical}. Darabi et al. proposed a variation of the derivative of the Swish-like activation in place of the STE mechanism for obtaining more effective back-propagation functions \cite{darabi2018bnn}. Lahoud et al. presented to use a smooth activation function at the beginning, and then gradually sharpened it to a binary representation equivalent to \texttt{sign} during the training \cite{lahoud2019self}. Hou et al. discussed loss-aware binarization, showcasing a proximal Newton algorithm with diagonal Hessian approximation that could directly minimize the loss with respect to the binary weights \cite{hou2016loss}. Observing that existing BNNs using real-valued latent weights to accumulate small update steps, Helwegen et al. viewed the latent weights as inertia, and introduced a BNN-specific optimizer called \emph{Binary Optimizer} (Bop) \cite{helwegen2019latent} for the training. Focusing on other training aspects, Tang et al. used special regularization items to encourage the latent floating-point variables approaching +1 and -1 during the training \cite{tang2017train}, Umuroglu et al. placed pooling before batch normalization and activation \cite{umuroglu2017finn}. Mishra et al. guided the training of BNNs through a well-trained, full-precision teacher network by adjusting the loss function. Additional BNN works could be found in the two surveys \cite{simons2019review, qin2020binary}. 

This work falls into the second category, aiming at enhancing BNN model capacity by enhancing the neuron's expressibility. BCNN uses dual bits for each complex binary neuron. Nevertheless, it is fundamentally different from 2-bit quantization; each 2 bits here embed a binary complex calculation logic, which, as shown later, is capable of extracting more expressive features. To ensure fairness, in BCNN, without changing the model structure, we proportionally reduce the number of channels to ensure consistent model size. 


\subsection{Complex Neural Network}

Complex numbers extend one dimensional real number line (i.e., -$\infty$ to $\infty$) to two dimensional complex plane by using the real axis and the imaginary axis. Although complex numbers do not exist in the real world, its unique properties and computing rules provide useful amendments to the representativeness of real numbers, especially when phase information is presented. For example, in physics, complex numbers are more suitable for representing waves, as the coefficients are complex after the Fourier transform; in neuroscience, neuronal rhythms, which are crucial for neuronal communication, are characterized by the firing rate and the phase, thus can be naturally expressed as complex numbers \cite{reichert2013neuronal}; in geoscience, PolSAR images \cite{gao2018enhanced, cao2019pixel} can offer much more comprehensive and robust information compared with pure SAR images. The scattering properties of PolSAR images can be naturally described by the complex-valued polarization scattering matrix, where the amplitude of each element corresponds to the back-scattering intensity of the electromagnetic wave from the target to the radar, and the phase corresponds to the distance between the sensor platform and the target. In biomedical science, being able to effectively handle phase information greatly facilitates MRI image reconstruction \cite{cole2020analysis, wang2020deepcomplexmri}.

Due to the richer representation and the need to process complex input signals with phases, there have long been efforts in constructing complex neural networks dating back to the 1990's \cite{georgiou1992complex, kim2003approximation, leung1991complex, kim2002fully}. The most recent one --- \emph{Deep Complex Networks} (DCN) was proposed by Trabelsi et al. \cite{trabelsi2017deep}, which formulated the building blocks of complex-valued deep neural networks include complex convolution, complex batch normalization, complex weight initialization strategy, etc. DCN takes into consideration the correlation between the real part and imaginary part of the complex inputs and weights, demonstrating its effectiveness on classification tasks, showing comparable or even superior performance than real-valued DNNs with only half of the real-valued network size.

This work is motivated by both DCN and BNN. We try to systematically integrate the two planes so that: (i) BCNN can show advanced accuracy compared to BNNs, with its richer representation through binary complex numbers; (ii) BCNN can drastically reduce the execution cost compared with DCNs, which is particularly attractive for embedded and edge utilization, where low cost, small size, low energy, and real-time response are usually demanding; (iii) Compared to BNNs, BCNN can naturally handle complex input signals, such as wave information directly from the sensors.


\section{Binary Complex Neural Network}
We present our binary complex neural network (BCNN) in this section. We first define a binary complex number and discuss its convolution process. We then present how to perform complex binarization and binary complex batch normalization. Finally, we propose a weight initialization strategy for BCNN.


\subsection{Binary Complex Number}

Similar to a complex number $z = x+iy$ that comprises a real part "$x$" and an imaginary part "$iy$", a \emph{binary complex number} is defined as $z^b = x^b+iy^b$ where $x^b$, $y^b$  $\in \{+1,-1\}$. Therefore, $z^b$ has four potential values: $\{-1-i$, $-1+i$, $1-i$, $1+i\}$, which can be encoded by two digital bits: the first one implies whether the real part is $-1$ or $+1$, while the second implies whether the imaginary part is $-i$ or $i$. 

For dot-product, if $z^b = x^b+iy^b$ is the binary complex input, $W^b = A^b+iB^b$ is the binary complex weight, then $h=c+id$ is the full-precision complex output. The bias is also a full-precision complex number, which is omitted in the following discussion for simplicity. Therefore, its dot product follows the complex calculation rules: 
\begin{equation}
h = c+id = (A^b*x^b-B^b*y^b) +i(B^b*x^b + A^b*y^b)
\end{equation}
where $x^b, y^b, A^b, B^b\in\{+1,-1\}$. Compared to BNN binary dot-product, a BCNN dot-product incorporates 4 binary dot-products and two extra real-valued additions. In the matrix notation, it is expressed as:
\[ \left[ \begin{array}{ccc}
c\\
d\\ \end{array}\right] = 
 \left[ \begin{array}{ccc}
A^b & -B^b\\
B^b & A^b\\ \end{array}\right] *
\left[ \begin{array}{ccc}
x^b\\
y^b\\ \end{array}\right]
\]

\subsection{Quadrant Binarization}
Binarization in BNN is the process of converting a full-precision real number into a binary number: $+1$ or $-1$, which is generally viewed as the non-linear activation function for BNNs.

Binarization can be performed in two approaches, known as \emph{deterministic} and \emph{stochastic} \cite{courbariaux2016binarized, hubara2016binarized}. The stochastic one can potentially offer a better accuracy, but at the expense of high implementation cost, whereas the deterministic one is merely a \texttt{sign} function, as shown below:
\begin{equation*}
  \text{Forward:} \quad 
  r^b = sign(r) = \begin{cases}
    +1 & r\geq 0 \\
    -1 &\text{otherwise}  \\
  \end{cases}       \\
\end{equation*}
Most BNN works adopt the low-cost deterministic binarization function. Since \texttt{sign} is non-differentiable at 0, and its gradient is always 0 otherwise,  direct back-propagation is infeasible. Prior works proposed the \emph{Straight-Through-Estimator} (STE) to do the back-propagation: 
\begin{equation*}
\text{Backward:} \quad \frac{\partial Loss}{\partial r} = \frac{\partial Loss}{\partial r^b} \mathbf{1}_{|r|<t_{clip}}
\end{equation*}
where $r$ is the full-precision real input. $r^b \in \{+1,-1\}$ is the binary output. $Loss$ is the value of the cost function. $t_{clip}$ is a clipping threshold, which typically sets to $1$. The gradient of \texttt{sign} function is simply set as an idientity function. The threshold is used to cancel the gradient, when the inputs are geting too large, which can assist in the optimization process.

\begin{figure}[!t]
\centering
\includegraphics[width=0.55\columnwidth]{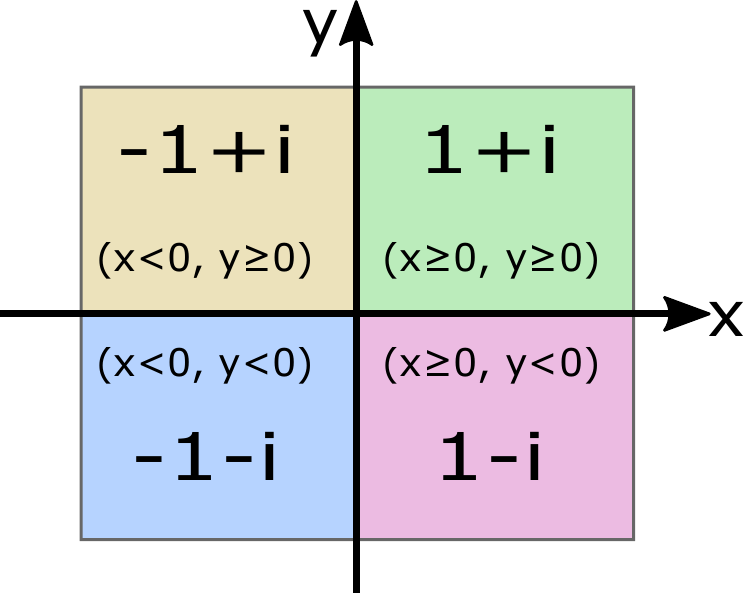} 
\caption{Quadrant binarization into a binary complex number.}
\label{fig:quandrant}
\end{figure}

The binarization to a binary complex number is to convert a complex number into a binary complex number (i.e., one of \{$1+i$, $1-i$, $-1+i$, $-1-i$\}). We propose quadrant binarization where the output is determined based on which quadrant the input complex number belongs to in the two-dimensional Cartesian system, as shown in Figure~\ref{fig:quandrant}. Mathematically, a complex plane is a geometric representation of the complex numbers settled by the real axis $x$ and the orthogonal imaginary axis $y$, where the two axes partition the plane into four quadrants, each bounded by two half-axes. Given four quadrants and four complex binary values, it is natural to link each quadrant with a complex binary value. The quadrant binarization is determined by the phase of a complex number, which is critical information in the complex signals.

This quadrant binarization essentially decouples the real part and the imaginary part so that both parts could be processed separately as an ordinary binarization. For the forward propagation, the binarization is as:
\begin{equation}
z^b = \text{sign}(x+iy) = \text{sign}(x) + i \text{sign}(y) = x^b + i y^b    
\label{eq:bcnn_bin}
\end{equation}
For the backward propagation, the gradient of the binarization is through two STEs, and applied over two independent full-precision latent variables $x$ and $y$: 
\begin{equation}
\frac{\partial Loss}{\partial z} = \frac{\partial Loss}{\partial x^b} \mathbf{1}_{|x|<t_{clip}} + i\frac{\partial Loss}{\partial y^b}\mathbf{1}_{|y|<t_{clip}} 
\end{equation}
This keeps the simplicity of the binarization process for efficient hardware implementation and memory storage. Note that to improve accuracy, existing works propose various variants of the binarization function, such as scaling factor \cite{rastegari2016xnor,zhou2016dorefa,tang2017train,lin2017towards,darabi2018bnn}, approximated \texttt{sign()} function \cite{darabi2018bnn,lahoud2019self}, etc. However, according to \cite{bethge2019binarydensenet}, no obvious accuracy improvement had been observed by adopting these strategies. Therefore, in this work, we use the original BNN binarization function as the baseline \cite{courbariaux2016binarized, hubara2016binarized}, which can achieve the best theoretic execution performance and model compression rate.

In BNNs, in addition to the 32x memory storage and bandwidth savings by adopting 1-bit of a neuron (compared to 32-bits per floating-point neuron), the computation efficiency gains from the approach on how bit dot-product is performed: each 32 or 64 bit dot-product in DNN can be accomplished by a single exclusive-nor (\texttt{xnor}) operation followed by a population count (\texttt{popc}) operation, leading to 10-16x speedups \cite{li2019bstc}:
\begin{equation*}
x*w \approx x^b * w^b = \text{popc}( \text{xnor}(x^b, w^b))    
\end{equation*}
A key requirement here is to conserve this property of being hardware friendly. Essentially in BCN,  
\begin{align}
\begin{split}
 z*W  &= (x+iy)*(A+iB) \\
 &\approx sign(x+iy) * sign(A+iB) \\
 &= (x^b + iy^b)*(A^b+iB^b) \\
&= (A^b*x^b-B^b*y^b) +i(B^b*x^b + A^b*y^b) 
\end{split}
\label{eq:bcn_dot_prod}
\end{align}
which implies that a BCNN dot-product can be computed by 4 BNN dot-products (i.e., \texttt{popc-xnor}) plus two full-precision additions. The 4 BNN dot-product can be operated in parallel at the same time, so the latency theoretically is close to one BNN dot-product.



In a BCNN convolution layer, within the input, weight and output tensors, we use the first half for the real part and the second half for the imaginary part. Specifically, if the input tensor has $M$ complex feature maps, it is equivalent to the  input has $2M$ real-valued feature maps, where the first $M$ represent the real components $x$ and the remaining $M$ represent the imaginary components $y$. The same case applies to the output tensor. Consequently, the weight tensor is in size $(N \times M \times k \times k) \times 2$ where the former half refers to the real part of the complex weight (i.e., $A$ in Eq.~\ref{eq:bcn_dot_prod}), and the latter half refers to the imaginary part (i.e., $B$ in Eq.~\ref{eq:bcn_dot_prod}).

\subsection{Complex Gaussian Batch Normalization}
\emph{Batch normalization} (BN) \cite{ioffe2015batch} has been proposed to accelerate convergence speed and contribute to better training accuracy. In real-value DNN, BN first normalizes the input so that the mean becomes zero and the variance becomes one. It then adjusts the normalized input by scaling through a learnable gain factor, and shifting through a learnable bias, as shown below: 
\begin{equation}
 BN(r) = \frac{r-\mu}{\sqrt{\sigma^2+\epsilon}}*\gamma+\beta  
\label{eq:dnn_bn}
\end{equation}
where $r$ is the input, $\mu$ is the mean of the batch, $\sigma$ is the variance of the batch, $\gamma$ is the learned scale, $\beta$ is the learned shift, $\epsilon$ is a tiny number for numerical stability. 

BN is important for DNNs, but is vital for BNNs. In addition to the normalization of the input, the learned gain and bias essentially increase the model capacity, or learning capability of a BNN layer. Without BN, the training of BNNs is even unlikely to converge.

Different from Eq~\ref{eq:dnn_bn}, standardizing complex input to normal complex distribution is much more complicated, because in addition to normalizing the mean and variance, Batch normalization in \emph{complex neural networks} needs to ensure equal variance of the real and imaginary components. In deep complex Network \cite{trabelsi2017deep}, the complex batch normalization is treated as a 2D whitening transformation -- scaling the complex input by the square root of their variance along the real and imaginary components. This is achieved by multiplying the 0-centered data by the inverse square root of the covariance matrix:
\begin{equation}
\tilde{z} = (V)^{-\frac{1}{2}}(z-E[z]) \qquad \qquad  \text{BN}(\tilde{z}) = \gamma \tilde{z} + \beta
\label{eq:cnn_bn}
\end{equation}
where $z$ is the complex input, $E[z]$ is the mean of $z$, $V$ is the $2 \times 2$ covariance matrix. The scaling parameter $\gamma$ is a $2 \times 2$ positive semi-definite matrix with three degrees of freedom ( $\gamma_{ri}$ and $\gamma_{ir}$ is the same). The shifting parameter $\beta$ is also a complex parameter. $\gamma_{ri}$, $\gamma_{ir}$, $\beta$ are initialized with 0; $\gamma_{rr}$ and $\gamma_{ii}$ are initialized with $1/\sqrt{2}$.

As shown in Eq~\ref{eq:cnn_bn}, complex batch normalization involves the computation of matrix inversion and matrix square-root, which is too costly for the hardware. Besides, directly adopting this complex batch normalization approach in BCNN leads to poor training accuracy, or even non-convergence, which is shown in Section IV. Consequently, we propose a novel batch normalization method called complex Gaussian batch normalization (CGBN), which is more efficient and light-weight.

Our objective is to normalize the input complex signal to a standard complex normal distribution ($CN$) \cite{goodman1963statistical, picinbono1996second}. The standard complex normal random variable, also known as standard complex Gaussian random variable, is a complex random variable $z$ whose real and imaginary parts are independent normally distributed random variables with mean equals to zero and variance equals to 1/2. In mathematical form, $z\sim CN(0,1)$ implies:
\begin{equation*}
R(z) \perp I(z)\ \mbox {and} \  R(z) \sim N(0,1/2) \ \mbox{and} \ I(z) \sim N(0,1/2)
\end{equation*}
Consequently, we can separately normalize the real part and imaginary part of the input complex signal to a normal distribution with zero mean and $1/2$ variance: 
\begin{equation}
\tilde{z} = (\frac{z_r-\mu_r}{\sqrt{2\sigma_r^2+\epsilon}}) + i (\frac{z_i-\mu_i}{\sqrt{2\sigma_i^2+\epsilon}}) 
\end{equation}
The scaling parameter and shifting parameter are learnable complex values, the complex Gaussian batch-normalization is as shown below:

\begin{align}
\begin{split}
\text{CGBN}(\tilde{z}) &=  \gamma* \tilde{z} +  \beta \\
  &=  \gamma_r \tilde{z}_r - \gamma_i \tilde{z}_i + \beta_r + i(\gamma_r \tilde{z}_i + \gamma_i \tilde{z}_r + \beta_i)
\label{eq:CGBN}
\end{split}
\end{align}

where both the scaling parameter $\gamma$ and shifting parameter $\beta$ are learned during the training. $\gamma$ is initialized as $\frac{1}{\sqrt{2}}+i\frac{1}{\sqrt{2}}$. $\beta$ is initialized as $0+i0$.

This complex Gaussian batch normalization significantly reduces the computation complexity compared to Eq~\ref{eq:cnn_bn} by avoiding the calculation of the inverse square-root of the covariance matrix. The complex Gaussian batch normalization leads to faster convergence speed and can converge in all of the models and datasets we have evaluated.

\subsection{Binary Complex Weight Initialization}
A proper weight initialization strategy can largely avoid  exploding or vanishing gradient problem during backpropagation and accelerate the convergence speed during network training. Usually, the weight initialization follows two rules: (i) the input and output have the same variance in the forward propagation; (ii) the gradient of input and output have the same variance in the backward propagation.

Two weight initialization strategies are broadly used for deep neural networks: \emph{Xavier} \cite{glorot2010understanding} and \emph{He} \cite{he2015delving}. Xavier is suitable for symmetric activation functions such as \texttt{tahn}, \texttt{softsign}, etc. The initial weight parameters follow a uniform distribution with zero mean and variance equals to $2/(fan\_{in} + fan\_{out})$. \emph{He} is specially designed for \texttt{ReLU} like activation functions. The variance of the initialization distribution is $2/fan\_{in}$ instead.

Following the \emph{Xavier} \cite{glorot2010understanding} and \emph{He} approach \cite{he2015delving}, the complex neural network \cite{trabelsi2017deep} derives the variance of the complex weight parameters. In the complex weight Initialization, a complex weight has a polar form:

\begin{equation}
W = |W|e^{i\theta} = R\{ W \} +i I\{ W\}
\end{equation}
The variance of $W$ is related to its amplitude not its phase, So the amplitude $|W|$ is set to follow the Rayleigh distribution, with the probability density function being: 
\begin{equation*}
 f(x,\sigma)=\frac{x}{\sigma^2}e^{-x^2/(2\sigma^2)}, x\ge0    
\end{equation*}
For Xavior \cite{glorot2010understanding}, to ensure $Var(W) = 2/(fan\_{in} + fan\_{out})$, then the parameter $\sigma=1/\sqrt{fan\_{in} + fan\_{out}}$. For He \cite{he2015delving}, to meet $Var(W) = 2/fan\_{in} $, set $\sigma = 1/\sqrt{fan\_{in}}$. The phase $\theta$ is unrelated to the variance, so it is initialized to follows the uniform distribution between $-\pi$ and $\pi$.

For BCNN, however, the complex weight initialization strategy does not work. After the quadrant binarization, the amplitude of binary complex weight is always $sqrt(2)$, which diminishes the efficiency of the original initialization strategy, which will be presented in our evaluation (Section~IV). 
Here, following the Xavier approach \cite{glorot2010understanding}, we derive our BCNN's weight initialization. As discussed in Section~III-A, in a BCNN layer $l$, the complex output $h_l = c_l+id_l$ is obtained by the convolution of the complex input $z_l = x_l+iy_l$ and the complex weight $W_l = A_l+iB_l$, the complex bias is ignored for simplification. The $f$ is the activation function, we have:
\begin{equation*}
c_l = x_l*A_l - y_l*B_l  \qquad  x_{l+1} = f(c_l)
\end{equation*}
\begin{equation*}
d_l = x_l*B_l + y_l*A_l  \qquad  y_{l+1} = f(d_l)    
\end{equation*}

The variance of real part and imaginary part can be written as:

$$Var[c_{l}] = fan\_ in*(Var[x_l]Var[A_l] + Var[y_l]Var[B_l]) $$
$$Var[d_{l}] = fan\_ in*(Var[x_l]Var[B_l] + Var[y_l]Var[A_l]) $$
If $C_{in}$ and $C_{out}$ are the channel size of the complex input and output, then $fan\_{in} = k^2 \times C_{in}$, $fan\_{out} = k^2 \times C_{out}$. For the Forward propagation, we ensure the real/imaginary part of input and output have the same
variance: $Var[c_l] = Var[x_l]$, $Var[d_l] = Var[y_l]$. At the same time, we assume the real part and the imaginary part of complex feature maps have the same variance: $Var[c_l] = Var[d_l]$, $Var[x_l] = Var[y_l]$, and for the complex weight as well: $Var[A_l] = Var[B_l]$, so we have:
\begin{equation*}
2 \times fan\_in \times Var[A_l] = 1    
\end{equation*}
\begin{equation*}
2 \times fan\_in \times Var[B_l] =1    
\end{equation*}
\begin{equation}
Var[A_l] = Var[B_l] =  \frac{1}{2 \times fan\_in}    
\label{eq:var_forward}
\end{equation}

for the backward propagation, the gradient of is computed as:
\small
\begin{equation*}
\frac{\partial Loss}{\partial x_l} = \tilde{A}\frac{\partial Loss}{\partial c_l} + \tilde{B}\frac{\partial Loss}{\partial d_l}  = f'(c_l) \tilde{A}\frac{\partial Loss}{\partial x_{l+1}} + f'(d_l) \tilde{B}\frac{\partial Loss}{\partial y_{l+1}}
\end{equation*}
\begin{equation*}
\frac{\partial Loss}{\partial y_l} = \tilde{A} \frac{\partial Loss}{\partial d_l} - \tilde{B} \frac{\partial Loss}{\partial c_l} = f'(d_l) \tilde{A} \frac{\partial Loss}{\partial y_{l+1}} - f'(c_l) \tilde{B} \frac{\partial Loss}{\partial x_{l+1}}    
\end{equation*}

\normalsize
The weight $A$ and $B$ here is a $C_{in}$-by-$k^2C_{out}$ matrix while the gradient of the weight $\tilde{A}$ and $\tilde{B}$is a $C_{out}$-by-$k^2C_{in}$ matrix. For input and output, the gradient of the real/imaginary part should have the same variance. With the assumption that for complex feature maps, the variance of the gradient for real part and imaginary part are the same, we have:
\begin{equation*}
 Var[\frac{\partial Loss}{\partial x_l}] = 2 \times fan\_out \times Var[{A}] \times Var[\frac{\partial Loss}{\partial x_{l+1}}]   
\end{equation*}
\begin{equation*}
 Var[\frac{\partial Loss}{\partial y_l}] = 2 \times fan\_out \times Var[{B}] \times Var[\frac{\partial Loss}{\partial y_{l+1}}]   
\end{equation*}
Therefore, for the backward pass, we have:
\begin{equation}
Var[A_l] = Var[B_l] =  \frac{1}{2 \times fan\_out}    
\label{eq:var_backward}
\end{equation}
With a compromise of Eq~\ref{eq:var_backward} and Eq~\ref{eq:var_backward}, we have:
\begin{equation}
Var[A_l] = Var[B_l] = \frac{1}{fan\_in + fan\_out}
\end{equation}
Therefore, in BCNN, we initialize the weight matrix by following the normal distribution with mean $\mu = 0.0$ and variance $\sigma = \sqrt{1/({fan\_in + fan\_out})}$.



\section{Evaluations}
In this section, we evaluate the performance of BCNN for image classification using three deep neural network models: \emph{NIN-Net}, \emph{ResNet18}, and \emph{ResNetE18} on two popular image classification datasets, \emph{CIFAR10} and \emph{ImageNet}. 

\subsection{Experimental Setup}

We use PyTorch to train all models. First we compare the BCNN and BNN with similar architecture and the same parameter size. Then, we evaluate and compare three different batch normalization and three weight initialization strategies for BCNN. 

\begin{figure}[!t]
\centering
\includegraphics[width=1.0\columnwidth]{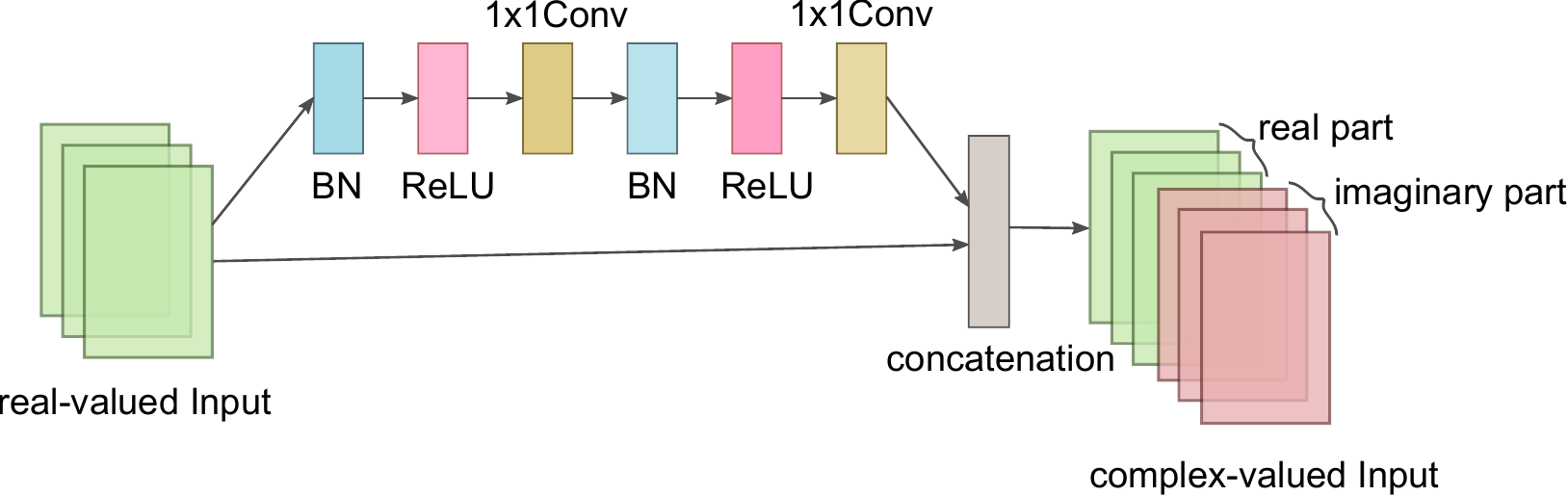} 
\caption{Structure to generate complex input from real-valued input.}
\label{fig:first_layer}
\end{figure}

\textbf{Complex-valued Input Data Generation:} As the raw data in Cifar10 and ImageNet datasets are real-valued, it is necessary to extend them to complex domain first. Our BCNN adopts a prior-art learning-based methodology proposed in \cite{trabelsi2017deep} to generate imaginary parts. As shown in Figure~\ref{fig:first_layer}, the imaginary parts are learned by a real-valued residual blocks, then the concatenation of raw real-valued data and the learned imaginary parts can serve as the complex-valued inputs. The two conv layers are both 1$\times$1 conv kernel, the channel of input and output are 3, so the real-valued residual block is lightweight in terms of both computation and storage.

\textbf{BCNN Model Configuration:} Usually in BNNs, the first and the last layer are full-precision. Using full-precision for the first layer is to conserve the maximum information flow from the input images. Using the full-precision for the last layer is to conserve the maximum state-space before the final output, which is in particular meaningful for large dataset like ImageNet.
For BCNN, we adopt the same strategy. Using full-precision complex convolution for the first layer. For the last layer, we use a full-precision real-valued layer by treating $M$ complex input as $2M$ real-valued input.

For fair comparison, all networks used in our evaluation of BCNN and BNN are with very similar architecture and the same model size. As BCNN uses complex-valued parameters, the model size of BCNN is about twice the size of BNNs with the same network configurations. Therefore, we tune the numbers of channels at each layer of BCNNs to approximately $1/\sqrt{2}$ of the ones in BNNs.  



\textbf{Network:} Three networks are used for evaluation: NIN-Net, ResNet18 and ResNetE18. 

(a) \textit{NIN-Net} consists of three stacked mlpconv layers followed by a spatial $2\times 2$ MAX-Pooling and a global Average-Pooling layer. For Cifar10 dataset, we use the original NIN-Net proposed in \cite{lin2013network}; for ImageNet dataset, we use the enhanced version of NIN-Net proposed in \cite{tang2017train}, which enlarges kernel sizes of the first four mlpconv layers from $1\times 1$ to $3\times 3$, and increase the output channels at the first two mlpconv layers from 96 to 128.

\begin{figure}[!htb]
\minipage{0.5\columnwidth}
\centering
\includegraphics[width=0.8\columnwidth]{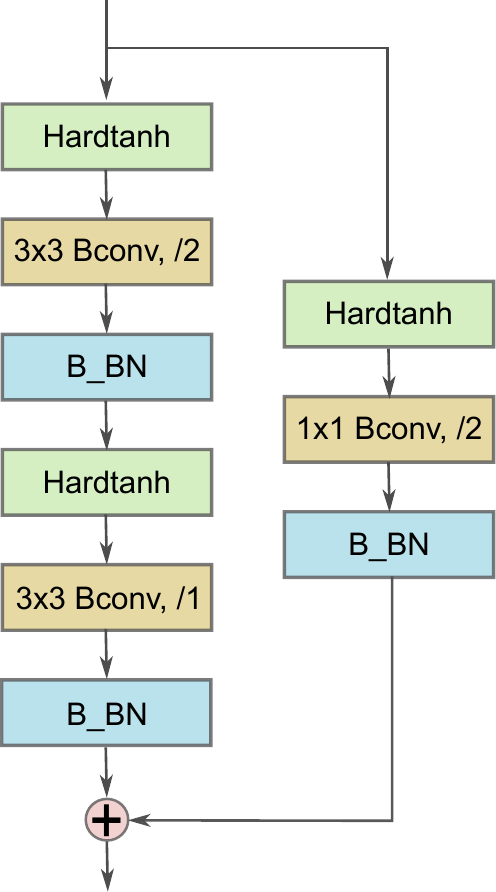} 
\caption{ResNet18}
\label{fig:resnet18}
\endminipage\hfill
\minipage{0.5\columnwidth}
\centering
\includegraphics[width=0.8\columnwidth]{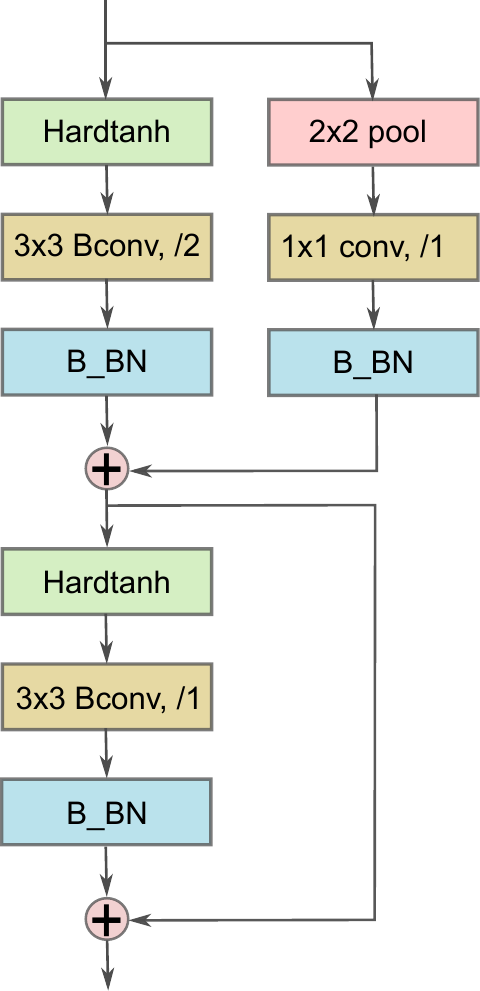} 
\caption{ResNetE18}
\label{fig:block}
\endminipage
\end{figure}

(b) \textit{ResNet18 and ResNetE18:} The block of ResNet18 and ResNetE18 are as shown in Figure.\ref{fig:resnet18} and Figure.\ref{fig:block} (the stride of first convolution is 2, when the stride is 1, the bypass will be identical). ResNetE is a modified version of ResNet, which is equipped with extra shortcuts and adopts full-precision down-sampling conv layer. With these modifications, ResNetE can preserve the information flow of the network better and process low-precision data, especially binary data, more efficiently.

\subsection{Result on CIFAR-10}

We first evaluate BCNN with NIN-net \cite{lin2013network} and ResNet18 \cite{he2016deep} on CIFAR-10 dataset. In training, Adam optimizer is adopted with the initial learning rate set as 5e-3. All models are trained for up to 300 epochs. We adjust the learning rates during training by multiplying them by 0.2 at the 80th, 150th, 200th, 240th, and 270th epochs respectively.

\begin{figure}[!htb]
\minipage{0.50\columnwidth}
\includegraphics[width=1\columnwidth]{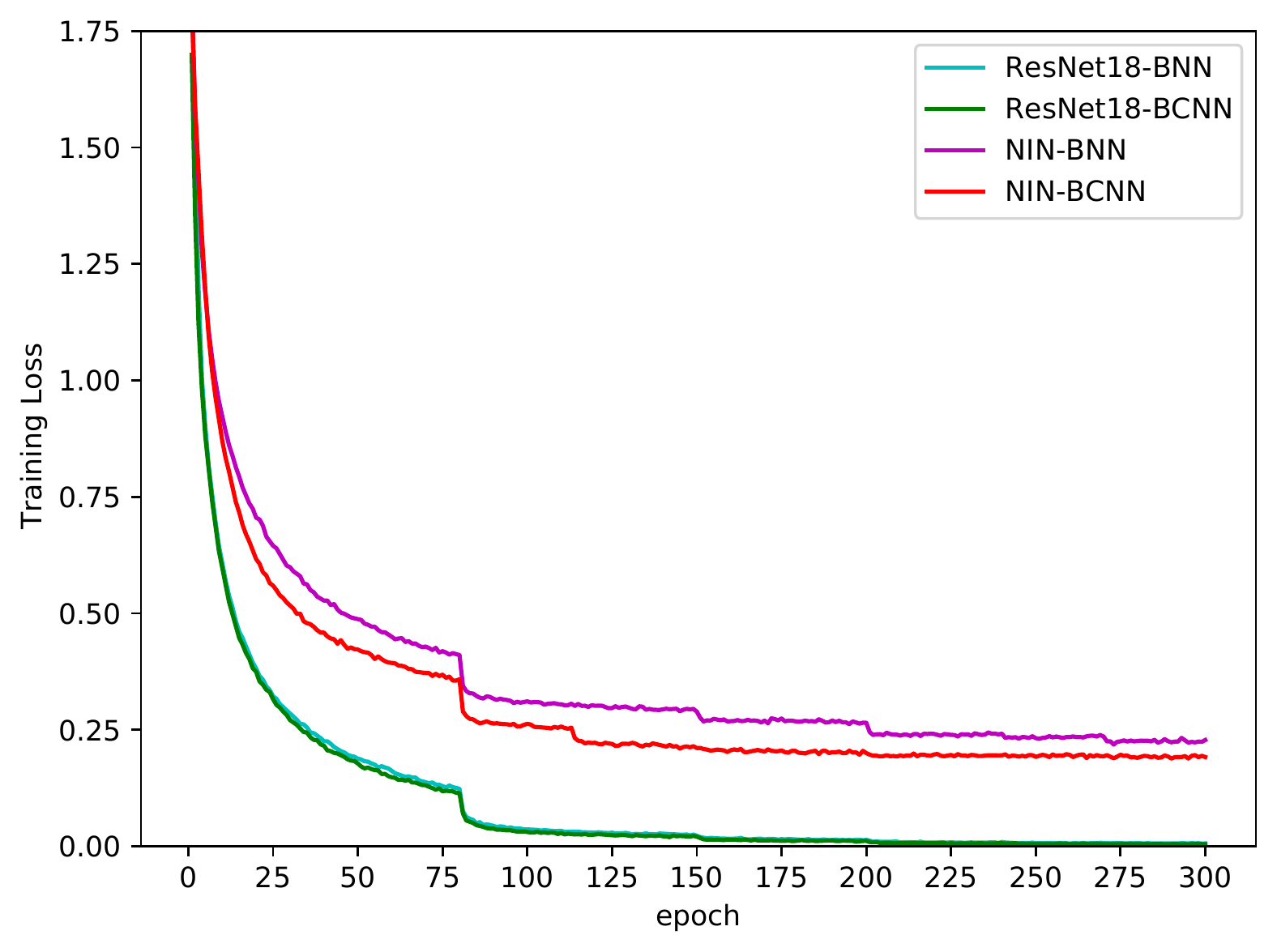} 
\caption{Training loss on Cifar10.}
\label{fig:3}
\endminipage\hfill
\minipage{0.50\columnwidth}
\includegraphics[width=1\columnwidth]{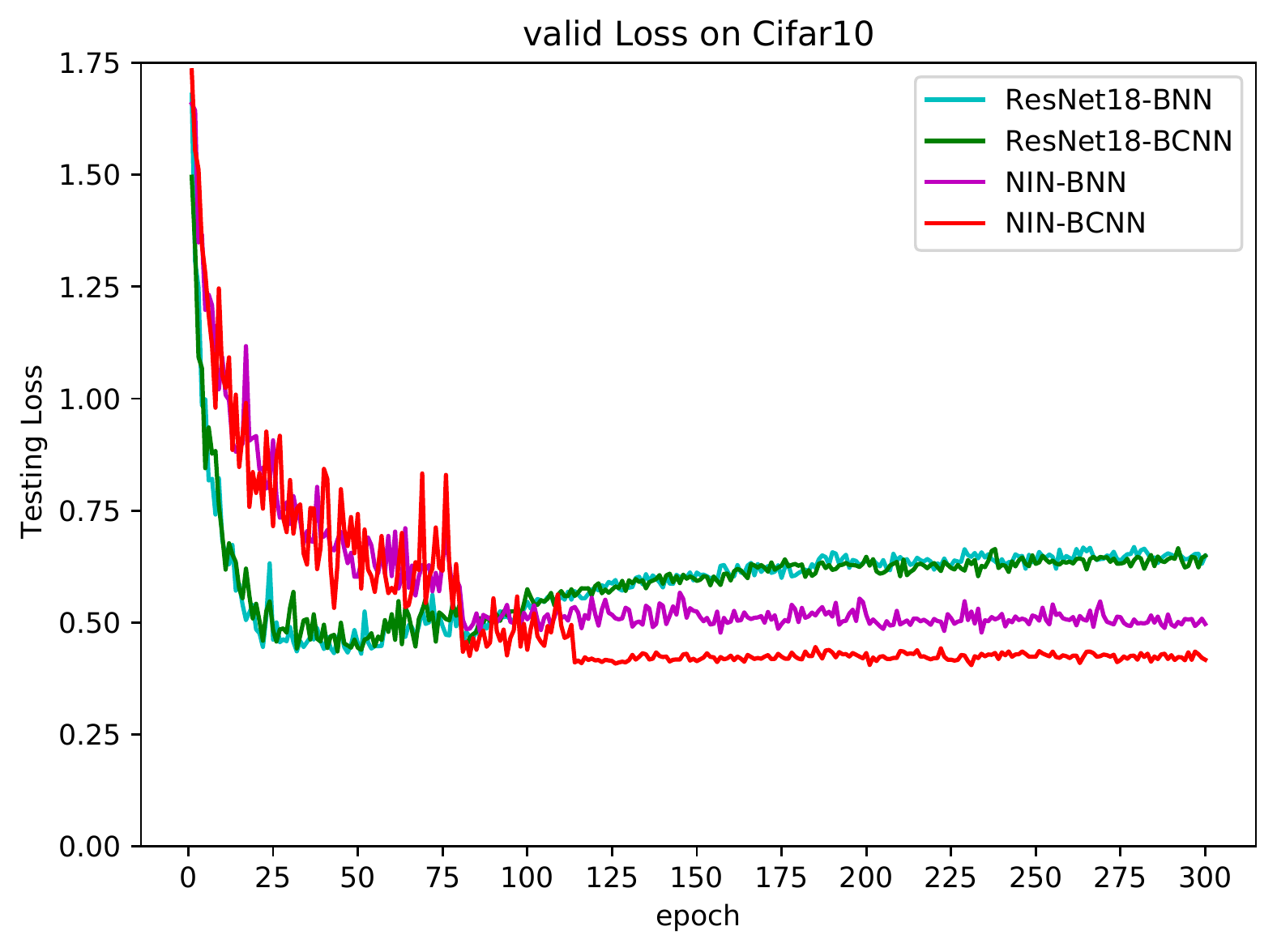} 
\caption{Testing loss on Cifar10.}
\label{fig:4}
\endminipage\hfill
\end{figure}

Table~\ref{Acc_cifar10} compares the accuracy of BCNNs, DNNs and BNNs. For NIN-Net and ResNet18, BCNN achieves 1.85\% and 0.52\% improvement on accuracy respectively over BNNs with the same model sizes. Figure.\ref{fig:3} and Figure.\ref{fig:4} show the variation of training loss and testing loss from epoch 0 to epoch 300.

\begin{table}[h]
\centering
\caption{Test accuracy on Cifar10.}
\label{Acc_cifar10}
\begin{tabular}{c c c c} 
\hline
\hline
Network & Type & Params & Top-1(\%) \\
\hline
NIN-Net & DNN & 3.69M & 89.64 \\
       & BNN & 0.191M & 85.77 \\
      & BCNN & 0.187M & \textbf{87.62} \\
\hline
ResNet18 & DNN & 42.63M & 93.02 \\
         & BNN & 1.39M & 90.67 \\
         & BCNN & 1.39M & \textbf{91.19} \\
\hline
\hline
\end{tabular}
\end{table}


\begin{table}[h]
\centering
\caption{Test accuracy with respect to batch normalization and weight initialization strategies on Cifar10. CGBN refers to Gaussian batch normalization proposed in this work. BN refers to normal batch normalization method \cite{ioffe2015batch}. CBN refers to the baseline batch normalization approach proposed in deep complex network \cite{trabelsi2017deep}. BCW refers to the binary complex weight initialization proposed in this work. Xavier refers to the Xavier weight initialization strategy \cite{glorot2010understanding}. Ray refers to the complex weight initialization approach proposed in deep complex network \cite{trabelsi2017deep}. NA means non-convergence.}
\label{diff_cifar10}
\begin{tabular}{c c c c} 
\hline
\hline
Network & BatchNorm & Weight\_init & Top-1(\%) \\
\hline
NIN-Net & CGBN  & BCW    & \textbf{87.62} \\
        & CGBN   & Xavier & 86.76 \\
        & CGBN  & Ray    &  86.07\\
        & CBN    & BCW    &  \textbf{87.86}          \\
        & BN   & BCW     &   87.38    \\
        & CBN    & Ray    & NA    \\
\hline
ResNet18 & CGBN  & BCW    & \textbf{91.19} \\
        &  CGBN   & Xavier & 90.56 \\
        &  CGBN  & Ray    & 89.92 \\
        & CBN    & BCW    &  \textbf{91.31}          \\
        & BN   & BCW    &  90.25     \\
        & CBN   & Ray    & NA  \\
\hline
\hline
\end{tabular}
\end{table}

In Table~\ref{diff_cifar10} , we show the impact of different batch normalization methodologies and weight initialization strategy on BCNN. We evaluate three different batch normalization. BN (batch normalization) is the most popular batch normalization that being used in real-valued neural networks. CBN (complex batch normalization) was proposed in deep complex network. CGBN (complex gaussian batch normalization) is proposed in this work for our BCNN.
Compared with the BN, the proposed CGBN improves the accuracy on both NIN-Net and ResNet18. For Cifar-10 dataset, our experimental results show that the CBN Batch Normalization can get higher accuracy compared to our CGBN. However, CBN is more computationally intensive and leads to non-convergence with large dataset, e.g. ImageNet. The detailed results on ImageNet will be given in the next section (NA in the table means non-convergence).

Different parameter initialization schemes also affect performance of BCNNs. As shown in Table~\ref{diff_cifar10} . BCW (binary complex weight initialization) is proposed for BCNN in this paper. Xavier is used for real-valued network, Ray was proposed for deep complex network. Results show the proposed initialization technique results in 1.55\% and 1.27\% accuracy improvement for NIN-Net and ResNet18 respectively.


\subsection{BCNNs on ImageNet}
In this section, we show the performance of BCNNs on ImageNet dataset. We use the standard pre-processing: all images are resized to $256 \times 256$ and randomly cropped to $224 \times 224$ for training. The validation dataset is with a single center crop.  We use ADAM optimizer in training and set the initial learning rate as 5e-3. This learning rate is gradually adjusted during training by being divided by 5 at epochs of 25, 35, 40 and 45. Each model is trained for 50 epochs. We use 3 models in our evaluation: NIN-E (expanded version of NIN-net as introduced in Section. IV(A)), ResNet18, and ResnetE18.

\begin{figure}[!htb]
\minipage{0.5\columnwidth}
\includegraphics[width=1\columnwidth]{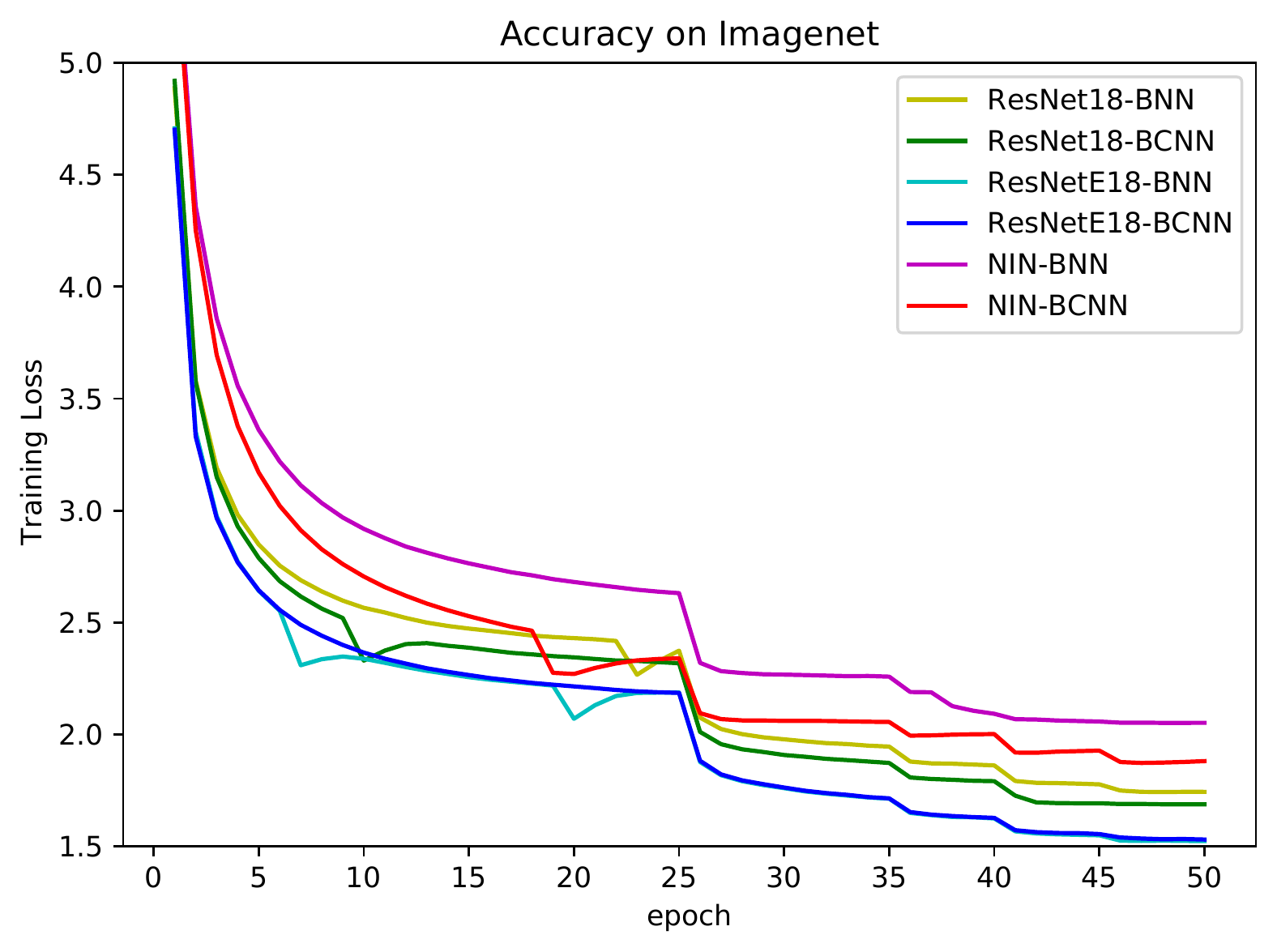} 
\caption{Training loss on ImageNet.}
\label{fig:5}
\endminipage\hfill
\minipage{0.5\columnwidth}
\includegraphics[width=1\columnwidth]{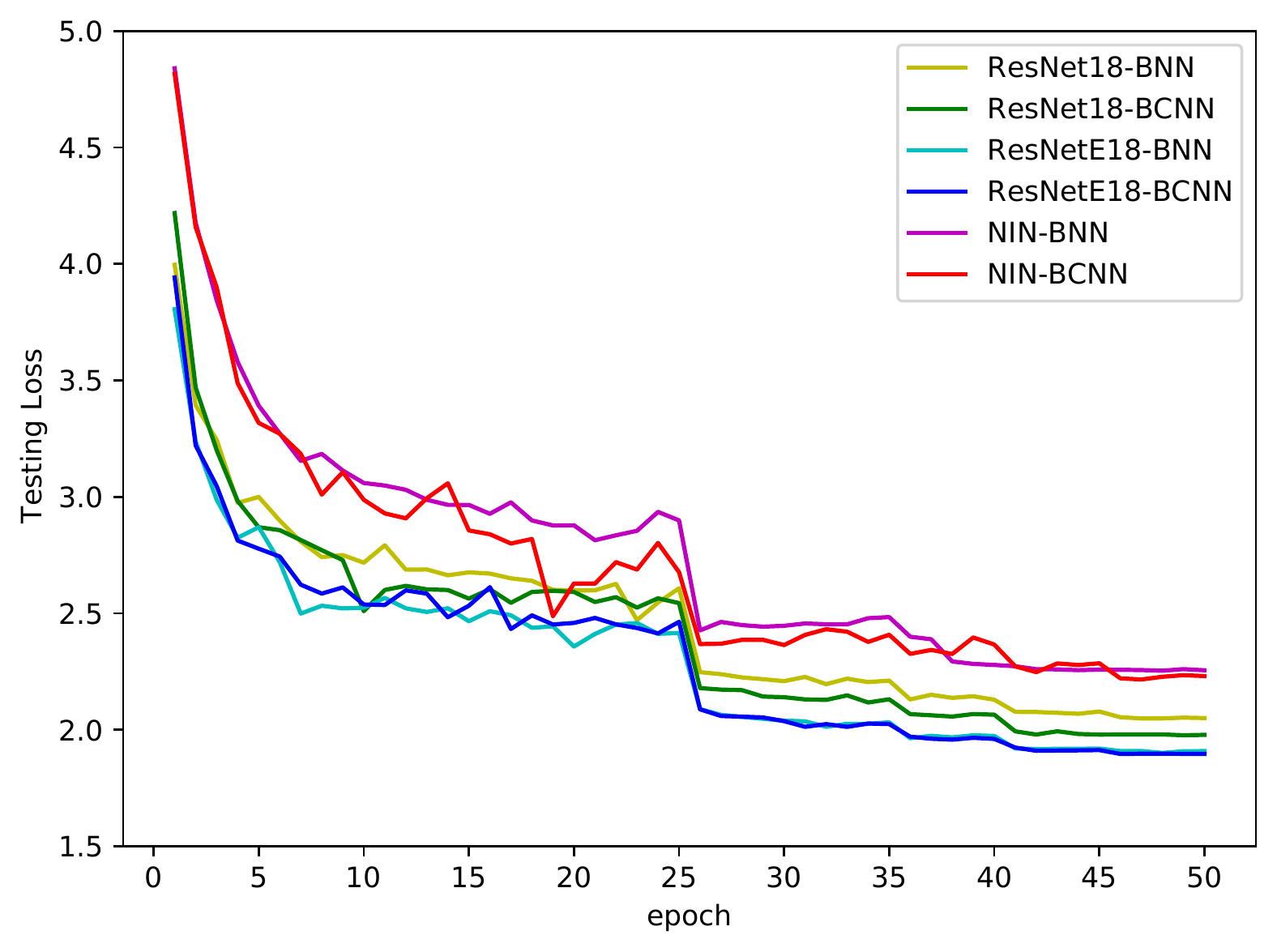} 
\caption{Testing loss on ImageNet.}
\label{fig:6}
\endminipage
\end{figure}

Table~\ref{acc_imag} compares the top-1 and top-5 accuracy of BCNNs, DNNs and BNNs. BCNN always has higher accuracy than BNN. For NIN-E, ResNet18 and ResNetE18, BCNN provides 0.8\%, 1.32\% and 0.17\% higher accuracy of top-1 than the ones of BNNs respectively. Figure.\ref{fig:5} and Figure.\ref{fig:6} show the changes of training and validation loss from epoch 0 to epoch 50.

\begin{table}[h]
\centering
\caption{Test accuracy on ImageNet.}
\label{acc_imag}
\begin{tabular}{c c c c c } 
\hline
\hline
Network & Type & Params & Top-1(\%) & Top-5(\%)\\
\hline
NIN-E & DNN & 28.96M & 57.09  & 79.34\\
        & BNN & 5.01M  & 50.018 & 73.936\\
       & BCNN & 5.0M & \textbf{51.08} & \textbf{74.738} \\
\hline
ResNet18 & DNN & 44.59M & 70.142  & 89.274 \\
         & BNN & 3.36M & 54.308  & 77.388 \\
         & BCNN & 3.36M & \textbf{55.598} & \textbf{78.708} \\
\hline
ResNetE18 & BNN & 4.0M & 57.332 & 79.85 \\
          & BCNN & 4.0M  & \textbf{57.652} & \textbf{80.016} \\
\hline
\hline
\end{tabular}
\end{table}

\begin{table}[h]
\centering
\caption{Test accuracy with respect to batch normalization and weight initialization strategies on ImageNet.}
\label{diff_imag}
\begin{tabular}{c c c c c } 
\hline
\hline
Network & BatchNorm & Weight\_init & Top-1(\%) & Top-5(\%)\\
\hline
NIN-E & CGBN  & BCW    & \textbf{51.08} & \textbf{74.738}\\
        & CGBN   & Xavier & 50.992  &  74.554\\
        & CGBN  & Ray    &  50.55  &  74.26\\
        & CBN    & BCW    &   \textbf{51.79}   &  \textbf{75.162}  \\
        & BN   & BCW     &  50.622  &   74.276    \\
        & CBN    & Ray    & NA      &\\
\hline
ResNet18 & CGBN  & BCW    & \textbf{55.598} & \textbf{78.708}\\
        &  CGBN   & Xavier & 54.98  & 78.238\\
        &  CGBN  & Ray    & NA \\
        & CBN    & BCW     &  NA    &    \\
        & BN   & BCW    &  54.91 &   78.148  \\
        & CBN   & Ray    & NA  & \\
\hline
ResNetE18 & CGBN  & BCW    & \textbf{57.652} & \textbf{80.016}\\
        &  CGBN   & Xavier & 57.250   & 79.976\\
        &  CGBN  & Ray    & NA \\
        & CBN    & BCW    &  NA   &    \\
        & BN   & BCW    &  57.642 &    80.188  \\
        & CBN   & Ray    & NA  & \\
\hline
\hline
\end{tabular}
\end{table}

The accuracy comparison of BCNNs with different Batch Norm techniques and weight initialization strategies are shown in Table~\ref{diff_imag}. 

BCNN with the proposed CGBN always provides higher accuracy than BN for all models. For NIN-E which is a relatively shallow network structure, CBN plus BCW have the highest accuracy. However, for the relativity deeper structures, e.g, ResNet18 and ResNetE18, CBN lead to non-converage during training. In contract, our proposed CGBN can still work efficiently.

We further evaluate the effects of different weight initialization strategies on ImageNet dataset. As listed in Table~\ref{diff_imag}, the proposed weight initialization BCW shows increased accuracy over Ray for NIN-Net. For ResNet18 and ResNetE18, the proposed BCW leads to faster convergence with higher accuracy than BNNs. As a comparison, BCNN with Ray cannot converge in our testing. 

Overall, we show that BCNN, together with the proposed batch normalization and weight initialization strategies, can achieve better training accuracy on some large datasets such as ImageNet. As the next step, we will seek efficient implementation of BCNN on various hardware platforms, including GPUs \cite{li2019bstc}, GPU Tensorcores \cite{li2020accelerating}, FPGAs \cite{geng2019lp, geng2020cqnn} and ASICs \cite{geng2019o3bnn, geng2020o3bnn}, for practical utilization in embedded systems and edge domains.

\section{Conclusion}
In this work we propose the binary complex neural network, which combines the advantages of both BNNs and complex neural networks. Compared to BNNs, it achieves enhanced training accuracy and is able to learn from complex data; compared to complex neural networks, it is much more computation efficient, which is in particular beneficial to terminal scenarios such as smart edges and smart sensors. Future work includes the demonstration of BCNN on complex datasets, the implementation of BCNN on embedded hardware devices, and its practical applications.


%




\ifCLASSOPTIONcaptionsoff
  \newpage
\fi



\bibliographystyle{IEEEtran}
\bibliography{references}
\end{document}